# Design and classification of dynamic multi-objective optimization problems

Alexandru-Adrian Tantar, Emilia Tantar and Pascal Bouvry
Computer Science and Communications Research Unit
University of Luxembourg
FirstName.LastName@uni.lu

November 10, 2010

## Abstract

In this work we provide a formal model for the different time-dependent components that can appear in dynamic multi-objective optimization problems, along with a classification of these components. Four main classes are identified, corresponding to the influence of the parameters, objective functions, previous states of the dynamic system and, last, environment changes, which in turn lead to online optimization problems. For illustration purposes, examples are provided for each class identified – by no means standing as the most representative ones or exhaustive in scope.

## 1 Introduction

The research trends in optimization for the past decade evolved from a focus on static single objective problems to complex cases where dynamic aspects and multiple objectives are being dealt with. While obvious that static approaches do not always model reality in a coherent manner, moving to dynamic scenarios and, what is more, while simultaneously optimizing more than one objective, proved to be extremely complex. We still need to understand what *dynamic* means, e.g. where is the limit between models defined as static snapshots at discrete moments in time and models subject to continuous evolution. Moreover we need to be able to define sound formal models and to understand what is the connection between dynamic changes in the input variables and the resulting effects in the objective space. A number of studies were conducted in the dynamic optimization literature, a first structural comprehensive view being given by the work of Branke [4]. Later papers, as detailed in the following, addressed particular aspects or extensions, introduced classifications of Pareto optimal set and front changes. Nonetheless, to the best of our knowledge, no formal framework was built to rigorously define the components needed to construct a dynamic problem formulation. We provide therefore in this paper a classification of the different dynamic elements that appear in the construction of such problems. For each class we provide simple and yet concise representative examples drawn from the literature in order to position each specific case and to offer a better understanding of the theoretical model.

Answering the question of what classes can be defined for dynamic multi-objective functions leads to (1) describing the combination of changes that appear in the Pareto front and set, as in Farina *et al.* [9], Mehnen *et al.* [14] or (2) delimiting the basic time-dependent components (e.g. in the decision space, environment). Having a clear classification is of primary importance not only for structuring notions and definitions but also for understanding the implications and the effects of each dynamic factor. On similar considerations, Weicker [16] outlined the role of classifications in establishing a strong foundation for systematic research in dynamic optimization. In the same work basic ideas for separating functions on translation and fitness rescaling basis are provided. An outline of the main criteria of classifcation adopted in the litterature is depicted in the following:

- studies dealing with periodicity, continuity or sparsity characteristics, environment time-dependent terms, behavior of specific algorithms, Yamasiki [17, 18];

- analyzing and characterizing changing morphology aspects, drifting landscapes, as discussed, for example, in Collard *et al.* [7] or De Jong *et al.* [8];

- function components that stand as a source of dynamism, as in Weicker [16] with parameter coordinate transforms and fitness rescaling;

- particular aspects of optimization algorithms and the role of parameters in problem generators, e.g. Branke [4], Yaochu [12].

A recent overview of the domain can be found in the work of Bu and Zheng [5], the first and last above perspectives being pointed out as of leading importance. An intense area of research over the past years was also in constructing functions for which the Pareto optimal front and set evolve with time, in either independent or simultaneous manner. A set of now *de facto* standard benchmarks resulted, generally constructed as extensions of classical static functions - refer to Farina *et al.* [9] or Mehnen *et al.* [14] for examples. Note that for most of these studies no time dependency is considered at an environment level (static definition), e.g. terms not specified inside the function itself, instance parameters. Moreover, the state of a dynamic system (function and environment) at a given moment in time is generally defined as or assumed to be independent of the previous states. At a different level, these aspects also lead to important aspects related to anticipation and prediction [10].

Finally, of specific interest for our work, in Mehnen *et al.* [14] four different types of multi-objective dynamic functions were identified where (I) the Pareto Optimal Set (POS) changes while the Pareto Optimal Front (POF) remains static, (II) both POS and POF vary, (III) only the POF changes and (IV) POS and POF are static. According to the specific type an arbitrary function belongs to and based on our current knowledge about respective effective paradigms, one should ideally be able to follow basic patterns when faced with a new problem. Nonetheless, this classification, although of undisputed importance, does not capture or does not describe the elements that turn a problem dynamic - where do these changes come from? For providing an answer to this concern we propose the following intuitive classification where the different basic points can be combined to construct higher complexity cases:

- **first order** - time-dependent *parameter evolution*: dynamic transform of the input parameters;

- **second order** - time-dependent *function evolution*: dynamic evolution of the objective functions values;

- **third order** - time-dependent *state dependency*: parameter or function state time-dependency, i.e. the parameters or the function is defined by taking into account not only the current moment in time but also the previous values or states;

- **fourth order** - time-dependent *environment*: parts of or the integral environment evolves with time.

A formal description and a more detailed discussion is given in the following sections along with simple examples. We consider that this classification comes to complete the one of Mehnen *et al.* [14] which does provide significant insights about the difficulty of the problem being dealt with. For an arbitrary function one may thus consider a more exact description where both the dynamic components and the induced effects at the Pareto boundary are specified, e.g. having a first order type I function.

The remainder of the article is divided in two sections. Section 2 starts by defining the component oriented formulation of dynamic multi-objective problems. The identified components subject to dynamic behavior are further used as criteria for classifying the existing approaches. The second criteria of classification is provided by the correlation between the states of the dynamic system inglobing the optimization problem. In Section 3 for each of the identified categories simple synthetic examples are provided in order to ilustrate the characteristics of each category. Section 4 concludes the article and present future research directions.

## 1.1 Background Notions and Notation

Let $F$ be an objective function, defined over $X$ (a decision space) and taking values in $Y$ (objective space), $F : X \to Y$, one may consider the $\min_{x \in X} F(x)$ minimization problem.

**Definition 1.** *Let $v, w \in \mathbb{R}^k$. We say that the vector $v$ is less than $w$ ($v <_p w$), if $v_i < w_i$ for all $i \in \{1, \ldots, n\}$. The relation $\leq_p$ is defined analogously.*

**Definition 2** (Dominance). *A point $y \in X$ is dominated by $x \in X$ $(x \prec y)$ if $F(x) \leq_p F(y)$ and if $\exists i \in \{1,...,k\}$ such that $f_i(x) < f_i(y)$. Otherwise $y$ is called non-dominated by $x$.*

**Definition 3.** *A point $x \in X$ is called a* Pareto *point if there is no $y \in X$ which dominates $x$. The set of all Pareto solutions forms the* Pareto set.

Furthermore, in order to model the dynamic behavior of the optimization problem, given $t$ a monotonically increasing value on a time period $[t_0, t_{end}] \in \mathbb{R}^+$ we have that:

$$\int_{t_0}^{t_{end}} F \mathrm{d}t = \left( \int_{t_0}^{t_{end}} f_1 \mathrm{d}t, \ \ldots \ , \int_{t_0}^{t_{end}} f_k \mathrm{d}t \right) \quad (1)$$

The following notation will be used as alternative to describe the dynamic formulation, for simplification purposes:

$$\int_{t_0}^{t_{end}} F \mathrm{d}t = \left( \int_{t_0}^{t_{end}} f_i \mathrm{d}t \right)_{1 \leq i \leq k} \quad (2)$$

## 2 Dynamic MO formulations

We define a multi-objective function vector $F_\sigma$ where $\sigma$ stands for an environment derived set of parameters. Note that this last term (the $\sigma$ term) is constant over time and generally omitted from notation, e.g. seed value, weights or constraints, except for online dynamic multi-objective formulations.

Let $H(F_\sigma, D, x, t)$ describe the behavior of a dynamic multi-objective optimization problem (system) having

- $F_\sigma$ - the multi-objective support function, $F_\sigma : X \to Y$, $F_\sigma(x) = [f_{\sigma,1}(x), \ \ldots \ , f_{\sigma,k}(x)]$;

- $D = [d_1, \ \ldots \ , d_k]$ a vector of time dependent functions, modeling the dynamic behavior of the time-changing component of the system;

- $x$ the set of variables/parameters;

- $t$ the time moment.

We denote the state of the system at time $t$ as described by the values of the state variables the parameters of $F_\sigma$ and $D$.

Based on the different types of dynamic transformations that can apply we depict in the following a classification for dynamic multi-objective problems, knowing that we consider the first order formulation as base formulation.

In this paper we concentrate our attention on the classification of the **dynamic multi-objective problems**, regardless of the existing evolutionary optimization techniques. The two criteria used in the classification are: (1) the component that models the dynamic behavior of the optimization problem and

| Dynamic component | Independent System States | Correlated System States |
|---|---|---|
| Parameters | $1^{st}$ order | $3^{rd}$ order |
| Functions | $2^{nd}$ order | $3^{rd}$ order |
| Environment | $4^{th}$ order | $4^{th}$ order |

Table 1: Classes of multi-objective dynamic optimization problems

(2) the dependency of the states of the dynamic optimization system at a given time moment $t$, $t \in [t_0, t_{end}]$ on the previous states of the system. A separation is made in the following where dynamic problems are modeled over a *static* support function subject to external dynamic factors that may act in the decision space, as part of the environment or at objective space level. Arguably one may consider that a first order dynamic problem, i.e. where time-dependent transformations are applied in the decision space, for example, can also be modeled as a second order problem with objective space time-dependent changes. Answering this concern requires first reviewing the following considerations:

- although academic examples fail to capture this aspect, the nature and the level at which dynamic factors act are imposed by the nature of the problem being dealt with and do not represent a modeling option – classifying a problem in one particular class is therefore subject to semantic considerations;

- real-life dynamic functions do not always have an explicit form (e.g. they can be black-box functions) and may only be approximated for static cases – one may hence be constrained to deal with time-dependent environment, input or output perturbations or transforms (external and not subject to an explicit control);

- assuming an explicit form or an approximation of the support function is given (*static* form), a decomposition follows where dynamic factors *are determined* by the transformations occuring over the input, environment or output spaces.

With respect to these considerations, we focus in the following on identifying the main dynamic components one has to deal with when addressing dynamic problems. Please note that complex combinations are possible where, for example, dynamic transforms are applied at both decision and objective space level at the same time (dynamic time-changing parameters and functions. While similar results may be obtained for different dynamic factors, e.g. the exact same displacement of the Pareto set or front, understanding the nature of the acting dynamic components can help in the construction of effective optimization algorithms.

All these classes can be further declined in the four types defined in [9], according to the Pareto front, Pareto set. In this article we will cover only the first 4 types, without giving an extensive view of online dynamic state/parameters/functions, in order to provide a base ground for the differentiating betwen the dynamic components behaviour, by means of simple/elementar test cases.

## 2.1 Dynamic parameter-time evolution

**Characteristics**: *dynamic transform applied on the input variables (decision space)*. *The external environment (described by $\sigma$) does not change with time.*
**Formulation**:
$$H(F_\sigma, D, x, t) = F_\sigma(D(x, t)),$$
where
$$F_\sigma(D(x,t)) = [\ f_1(d_1(x,t)),\ \ldots,f_k(d_k(x,t))\ ]$$

In this case the dynamic evolution of the system in time is made through a transformation (noise function) applied on the input variables. The support function remains unchainged. The associated optimization problem, becomes thus:
$$\min \int_{t_0}^{t_{end}} F_\sigma(D(\mathbf{x}(t), t)) \mathrm{dt},$$
where $[t_0, t_{end}]$ depicts the time interval on which the behavior of the dynamical system is observed.

$$\min_{\mathbf{x}(t)} \left\{ \left( \int_{t_0}^{t^{\text{end}}} f_i(\ d_i(\mathbf{x}(t),t)\ )\ \mathrm{d}t \right)_{1 \leq i \leq k} \right\}$$

## 2.2 Dynamic function evolution

**Characteristics**: *dynamic transform applied on the support function (affecting the behavior in the objective space), e.g. superposed noise evolving with time. The external environment (described by $\sigma$) does not change with time.*

*Note*: In this case, as well as in the previous one, the succesive states of the system from time $t_0$ to time $t_{end}$ are independent, the same transformation function $D$ and same support vector of functions $F_\sigma$ being applied at every time moment.

**Formulation**:
$$H(F_\sigma, D, x, t) = D(F_\sigma, x, t)$$
with
$$D(F_\sigma, x, t) = (d_i(F_\sigma, x, t))_{1 \leq i \leq k}$$
classically defined on the respective objective function:
$$D(F_\sigma, x, t) = (\ d_i(f_{\sigma,i}, x, t))_{1 \leq i \leq k}$$

The time dependence of the input variables/parameters $x$, where the values of $x$ are scaled with time, is translated in optimization problem formulation as

$$\min_{\mathbf{x}(t)} \int_{t_0}^{t_{end}} D(F_\sigma, \mathbf{x}(t), t) \mathrm{d}t.$$

or more explicitely

$$\min_{\mathbf{x}(t)} \left\{ \left( \int_{t_0}^{t^{\text{end}}} d_i(\, F_\sigma, \mathbf{x}(t), t\,)\,) \, \mathrm{d}t \right)_{1 \leq i \leq k} \right\}$$

It should be noted that we can distinguish also according to the number of functions that are affected.

## 2.3 State-dependency

**Characteristics**: *the dynamic transform at time $t$, takes into account values obtained by the support functions/variables at $j$ previous time moments, having $1 \leq j \leq t$. The external environment (described by $\sigma$) does not change with time.*

**General formulation**: Let us consider that the analytical form of $H$ at a given time moment $t$ fuly describes the state of the dynamic system, defined through

$$H(F_\sigma, T^{[t-j,t]}, x, t)$$

where the associated dynamic multi-objective optimization problem can be formulated as

$$\min_{\mathbf{x}(t)} \int_{t_0}^{t_{end}} H(F_\sigma, T^{[t-j,t]}, x(t), t) \mathrm{d}t$$

As opposite to the previous two cases were a *static dependecy function $D$* was defined for each state, we consider for this class that the values of $H$ at time $t$ depend on the values/formulations of the parameters or base functions at previous states of the system, given by a trasformation function $T^{[t-j,t]}$.

### 2.3.1 State-parameter-dependency

**Characteristics**: *The current state of the optimization system is expressed analytically in function of the past states from $t - j$ to $t$. In this first case of state-parameter dependency, the values of the set of parameters $x$ at time $t$ depend of the values of the $x$ parameters at the time $t - j$ to $t$.*
**Formulation**

$$H(F_\sigma, T^{[t-j,t]}, x, t) = F_\sigma(T^{[t-j,t]}(x))$$

The optimization problem becomes thus,

$$\min_{\mathbf{x}(t)} \left\{ \left( \int_{t_0}^{t_{end}} f_{\sigma,i}(\ T^{[t-j,t]}(x(t)),\ t)\ \mathrm{d}t \right)_{1 \leq i \leq k} \right\}.$$

*Note*: For the case were $x$ at the time moment $t$ depends only on the previous state, the application of the transformation function $T$, provides $F_\sigma(T^{[t-1,t]}(x))$.

### 2.3.2 State-Function-dependency

**Characteristics**: *The value of the objective function depends on the previous values of the base function $F_\sigma$ on a given time interval $[t-j,t]$. The optimal solution is defined on the previous values of the function $F$.*

**Formulation**

$$H(F_\sigma, T^{[t-j,t]}, x, t) = T^{[t-j,t]}(F_\sigma(x))$$

and the optimization problem becomes

$$\min_{\mathbf{x}(t)} \int_{t_0}^{t_{end}} T^{[t-j,t]}(F_\sigma(\mathbf{x}(t)))\mathrm{d}t.$$

## 2.4 Online dynamic multi-objective optimization

While the above formulations stands for dynamic cases with a static environment $\sigma$ being fixed, the online class englobes the case of dynamically changing environments, where $\sigma$ varies over time.

**Characteristics**: *the environment $\sigma$ changes dynamically with time.*

**Formulation**

The formulation of online dynamic multi-objective problems can be modeled as

$$H(F_\sigma, D, x, t) = F_{D(\sigma,t)}(x, t)$$

All the four previous classes ca be also extended to the case of dynamic environments, online dynamic multi-objective

Different classes can be designated here, with functions including time-evolving factors or enclosing random variables, functions depending on past states, etc. For the general case and assuming a minimization context, the goal is to identify a sequence of solutions $x(t)$, with $t \in [0, t^{end}]$ leading to the following:

$$\min_{\mathbf{x}(t)} \int_{t_0}^{t_{end}} F_{D(\sigma,t)}(\mathbf{x}(t), t)\ \mathrm{d}t$$

$$\min_{\mathbf{x}(t)} \left\{ \left( \int_{t_0}^{t_{end}} f_{D(\sigma,t),i}(\mathbf{x}(t), t)\ \mathrm{d}t \right)_{1 \leq i \leq k} \right\}$$

# 3 Dynamic multi-objective optimization: Test problems

The problems tackled so-far in the litterature of dynamic multi-objective optimization include:

- Moving Peaks Problem;
- Synthetic Dynamic Problems;
- Dynamic Sphere problem extensions (e.g. FDA).

Some references on applications of dynamic optimization problems treated by means of multi-objective forumlations can be found at [15].

### First order problem example

We illustrate the first order class by means of an extension of the dynamic unimodal sphere problem, by using the DSW model described in [14]. We model the DSW functions by using:

$$F(D(x,t)) = (f_1(d_1(x,t)),\ f_2(d_2(x,t)))$$

$$f_j(x) = \sum_{i=1}^{n} x_i^2,\ j = 1,2$$

where

$$d_1(x,t) = (a_{11}x_1 + a_{12}x_1 - b_1 \cdot G(t), x_2,\ \ldots,x_n)$$
$$d_2(x,t) = (a_{21}x_1 + a_{22}x_1 - b_1 \cdot G(t) - 2, x_2,\ \ldots,x_n)$$

and $G(t): \mathbb{R} \to \mathbb{R}$, $G(t) := t(\tau) \cdot s$, with $t(\tau) = \left\lfloor \frac{\tau}{\tau_{t_{end}}} \right\rfloor$.

By varying the values of the parameters $a_{11}, a_{12}, a_{21}, a_{22}, b_1, b_2$ three classes of problems can be derived, as depicted in [14].

### Second order problem examples

[**General formulations**]Different general formulations may apply, take for example:

- $D(F_\sigma, x, t) = D(F_\sigma(x,t))$,
- $D(F_\sigma, x, t) = F_\sigma(x) + D(x,t)$, where $D(x,t)$ can represent a Gaussian noise function, changing with time.

[**Function-Dynamic DTLZ2**] A classic instance of this class is represented by the dynamic variant of the DTLZ2 function [9], in its original form (static case) defined as follows:

$$F_\sigma(x) = [f_{\sigma,1}(x), f_{\sigma,2}(x), \ldots, f_{\sigma,k}(x)]$$

$f_1(x) = (1+g)cos(x_1\frac{\pi}{2})cos(x_2\frac{\pi}{2})\ldots cos(x_{k-2}\frac{\pi}{2})cos(x_{k-1}\frac{\pi}{2})$
$f_2(x) = (1+g)cos(x_1\frac{\pi}{2})cos(x_2\frac{\pi}{2})\ldots cos(x_{k-2}\frac{\pi}{2})sin(x_{k-1}\frac{\pi}{2})$
$f_3(x) = (1+g)cos(x_1\frac{\pi}{2})cos(x_2\frac{\pi}{2})\ldots sin(x_{k-2}\frac{\pi}{2})$
$\ldots$
$f_k(x) = (1+g)sin(x_1\frac{\pi}{2})$

$$0 \leq x_i \leq 1, for\ i = 1, 2, \ldots, n$$

$$g(\mathbf{x}_k) = \sum_{x_i \in \mathbf{x}_k} (x_i - 0.5)^2$$

with $\mathbf{x}_k$ a decision vector, $|\mathbf{x}_k| = p$, i.e. $p$ variables required for the $g(\mathbf{x}_k)$ function, given a total of $p+k-1$ variables (where $k$ is the number of objectives). In order to introduce the dynamic case, we proceed by first constructing the support function $F(x) = [f_1(x), f_2(x), \ldots, f_k(x)]$, with $f_i(x), 1 \leq i \leq k$ defined as:

$$f_i(x) = \begin{cases} \prod_{j=1}^{k-i} cos\left(x_j\frac{\pi}{2}\right), & i = 1 \\ sin\left(x_{k-i+1}\frac{\pi}{2}\right) \prod_{j=1}^{k-i} cos\left(x_j\frac{\pi}{2}\right), & 1 < i < k \\ sin\left(x_1\frac{\pi}{2}\right), & i = k \end{cases}$$

$$0 \leq x_i \leq 1, for\ i = 1, 2, \ldots, n$$

The dynamic form of the function is obtained by considering:

$$D(F_\sigma, x, t) = D((f_{\sigma,1}, \ldots, f_{\sigma,k}), x, t) =$$

$$[\ d_1((f_1, \ldots, f_k), x, t),\ \ldots\ ,\ d_k((f_1, \ldots, f_k), x, t)\ ]$$

where the environment $\sigma$ is constant over time, and omitted for simplicity purposes in the reaminder of the example and

$$d_i((f_1, \ldots, f_k), x, t) = g(\mathbf{x}_k, t) f_i(x)$$

$$g(\mathbf{x}_k, t) = G(t) + \sum_{x_i \in \mathbf{x}_k} (x_i - G(t))^2, 0 \leq G(t) \leq 1$$

Additional examples can also be found in [9, 14].

### Third order problem example

For the correlated states model we base our exemplification on the **multi-objective moving peaks** problem. The moving peaks problem was proposed by Branke [6] and it is defined on a landscape formed of $m$ peaks, each of them described through three parameters: height $h$, width $w$ and location $p$.

The objective function proposed by Branke is given by:

$$F(x,t) = \max \left\{ B(x), \max_{1 \leq i \leq m} \{P(x, h_i(t), w_i(t), p_i(t))\} \right\} \qquad (3)$$

The state-dependency can be obtained by noise added at each generational step to the values of the parameters. For the state-function dependency one can consider the fact that the objective functions are not computed on the base landscape $B(x)$, but reather on the landscape obtained at the previous step. The second objective function does not necesarily has to be dynamic, so we can consider that the second objective function formulation follow the same formulations as in [13], previously described.

### Fourth order problem example

An example of online dynamic multi-objective problem involving a real-life application was depicted in [11] for a dynamic hospital resource management. Other examples of online dynamic optimization problems involving a single objective optimization modeling were depicted in the work of Bosman [2, 3], for delivery pick-up optimization problems.

The MNK-landscape problem can be extended as to fit the dynamic fourth order class by changing the tables providing the values of the interaction at every time moment. In order to construct the multi-objective model used as vector of support functions, the formulation that can be considered is the MNK-lansdcape variant proposed by Aguirre and Tanaka [1]

## 4 Conclusions

Through this paper we aim at providing a formal framework useful in positioning the dynamic multi-objective problems that occur and outline the components that induce the dynamic behavior of the problems. For each class at least one example is provided in order to ilustrate how existing synthetic problems can be situated. The provided examples are by no means considered as the most representative for the given category, they are just included for illustration purposes.

# References

[1] H. Aguirre and K. Tanaka. Insights on properties of multiobjective mnk-landscapes. In *Congress on Evolutionary Computation (CEC 2004)*, volume 1, pages 196 – 203, june 2004.

[2] P. A. N. Bosman. Learning, anticipation and time-deception in evolutionary online dynamic optimization. In *GECCO Workshops*, pages 39–47, 2005.

[3] P. A. N. Bosman and H. L. Poutré. Learning and anticipation in online dynamic optimization with evolutionary algorithms: the stochastic case. In *GECCO*, pages 1165–1172, 2007.

[4] J. Branke. *Evolutionary Optimization in Dynamic Environments*. Kluwer Academic Publishers, Norwell, MA, USA, 2001.

[5] Z. Bu and B. Zheng. Perspectives in dynamic optimization evolutionary algorithm. In Z. Cai, C. Hu, Z. Kang, and Y. Liu, editors, *Advances in Computation and Intelligence*, volume 6382 of *Lecture Notes in Computer Science*, pages 338–348. Springer Berlin / Heidelberg, 2010.

[6] L. T. Bui, J. Branke, and H. A. Abbass. Multiobjective optimization for dynamic environments. In *Congress on Evolutionary Computation*, pages 2349–2356, 2005.

[7] P. Collard, C. Escazut, and A. Gaspar. An evolutionary approach for time dependent optimization. *International Journal on Artificial Intelligence Tools*, 6(4):665–695, 1997.

[8] K. De Jong. Evolving in a changing world. In Z. Ras and A. Skowron, editors, *Foundations of Intelligent Systems*, volume 1609 of *Lecture Notes in Computer Science*, pages 512–519. Springer Berlin / Heidelberg, 1999.

[9] M. Farina, K. Deb, and P. Amato. Dynamic multiobjective optimization problems: test cases, approximations, and applications. *IEEE Trans. Evolutionary Computation*, 8(5):425–442, 2004.

[10] I. Hatzakis and D. Wallace. Dynamic multi-objective optimization with evolutionary algorithms: a forward-looking approach. In *Proceedings of the 8th annual conference on Genetic and evolutionary computation (GECCO'06)*, pages 1201–1208, New York, NY, USA, 2006. ACM.

[11] A. K. Hutzschenreuter, P. A. N. Bosman, and H. L. Poutré. Evolutionary multiobjective optimization for dynamic hospital resource management. In *Evolutionary MultiCriterion Optimization 5th International Conference (EMO 2009)*, volume 5467 of *Lecture Notes in Computer Science*, pages 320–334. Springer, 2009.

[12] Y. Jin and J. Branke. Evolutionary optimization in uncertain environments-a survey. *IEEE Trans. Evolutionary Computation*, 9(3):303–317, 2005.

[13] X. Li, J. Branke, and M. Kirley. On performance metrics and particle swarm methods for dynamic multiobjective optimization problems. In *IEEE Congress on Evolutionary Computation (CEC 2007)*, pages 576 – 583, September 2007.

[14] J. Mehnen, T. Wagner, and G. Rudolph. Evolutionary optimization of dynamic multi-objective test functions. In *Proceedings of the Second Italian Workshop on Evolutionary Computation (GSICE2), CD-ROM proceedings*, September 2006.

[15] Models of Decision and Optimization (MODO) Research Group. Intelligent strategies in uncertain and dynamic environments. `http://dynamic-optimization.org`, January 2011.

[16] K. Weicker. Performance measures for dynamic environments. In *Parallel Problem Solving from Nature (PPSN VII)*, volume 2439 of *Lecture Notes in Computer Science*, pages 64–73. Springer Berlin, Heidelberg, 2002.

[17] K. Yamasaki. Dynamic pareto optimum ga against the changing environments. In *2001 Genetic and Evolutionary Computation Conference. Workshop Program*, pages 47–50, San Francisco, California, July 2001.

[18] K. Yamasaki, K. Kitakaze, and M. Sekiguchi. Dynamic optimization by evolutionary algorithms applied to financial time series. In *Proceedings of the 2002 Congres on Evolutionary Computation (CEC '02)*, volume 02, pages 2017–2022, Washington, DC, USA, 2002. IEEE Computer Society.